\documentstyle[nlpia,macros]{article}

 \include{epsf}

 \title{Improving Tagging Accuracy by Using Voting Taggers}
 \author{{\bf Llu\'{\i}s M\`arquez}, {\bf Llu\'{\i}s Padr\'o} and {\bf Horacio Rodr\'{\i}guez}\\
 Dep. LSI. Technical University of Catalonia\\
 c/ Jordi Girona 1-3. 
 08034 Barcelona\\
 {\tt \{lluism,padro,horacio\}@lsi.upc.es}}

\begin{document}

\maketitle
\bibliographystyle{nlpia}

\begin{abstract}
We present a bootstrapping method to develop an annotated corpus,
which is specially useful for languages with few available resources.
The method is being applied to develop a corpus of Spanish of over
5Mw. The method consists on taking advantage of the collaboration 
of two different POS taggers. The cases in which both taggers agree 
present a higher accuracy and are used to retrain the taggers.
\end{abstract}

\section{Introduction}
\label{seccio-introduccio}

Usual automatic tagging algorithms involve a process of 
acquisition (or learning) of a statistical language model
from a previously tagged training corpus (supervised 
learning). The statistical models contain lots of parameters 
that have to be reliably estimated from the corpus, so the 
sparseness of the training data is a severe problem.

  When a new annotated corpus for a language with a reduced
amount of available linguistic resources is developed, 
this issue becomes even more important, since no training 
corpora are available and the manual tagging of a big enough
training corpus is very expensive, both in time and human 
labour. If costly human labour is to be avoided, the accuracy 
of automatic systems has to be as high as possible, 
even starting with relatively small manually tagged 
training sets. 

  In the case of English, existing resources are usually enough, thus
existing work on developing corpora does not rely much in bootstrapping,
although re--estimation procedures are widely used to improve tagger
accuracies, specially when limited disambiguated material is 
available~\cite{church88,briscoe94,elworthy94}. We find automatically tagged
corpora which are hand corrected {\sl a posteriori} \cite{marcus93}, and fully
automatic disambiguation procedures \cite{leech94,jarvinen94}

  Bootstrapping is one of the methods that can be used 
to improve the performance of statistical taggers when only 
small training sets are available. 
  The bootstrapping procedure starts by using a small hand-tagged 
portion of the corpus as an initial training set. Then, the tagger 
is used to disambiguate further material, which is incorporated to the 
training set and used to retrain the tagger.
Since the retraining corpus can be much larger than the initial 
training corpus we expect to better estimate (or learn) the 
statistical parameters of the tagging model and to obtain a 
more accurate tagger.
Of course, this procedure can be iterated leading, hopefully, to 
progressively better language models and more precise taggers.
The procedure ends when no more improvement is achieved.

  As stated above, the bootstrapping refining process is completely 
automatic. However each step of training corpus enlargement and 
enrichment could involve a certain amount of manual revision and 
correction. In this way the process would be semi--automatic.

  The main problem of this approach is that the retraining material
contains errors (because it has been tagged with a still poor tagger) and that 
this introduced noise could be very harmful for the learning procedure
of the tagger.
Depending on the amount of noise and on the robustness of the 
tagging algorithm, the refining iteration could lead to no improvement 
or even to a degradation of the performance of the initial tagger.

  Recent studies~\cite{padro98b} point that the noise in training and 
test corpora are crucial not only for the right evaluation of an NLP 
system, but also for its appropriate training to get an optimal 
performance. So, keeping a low error rate in retraining material 
becomes an essential point if we want to guarantee the validity of the 
bootstrapping approach.

  In this paper we show that it is possible to take advantage of
the collaboration between two (or more) different taggers in a 
bootstrapping process, by observing that in the cases in which 
both taggers propose the same tag present a much higher precision
than any of them separately and that these coincidence cases    
represent a coverage of more than 95\%. 
Then, the corpus to retrain the taggers is built, at each step, 
on the basis of this intersection corpus, keeping fairly low 
error rates and leading to better language models and more 
precise taggers.

  In addition, it is clear that the combination of taggers can 
be used to get a high recall tagger, which proposes an unique 
tag for most words and two tags when both taggers disagree.
Depending on the user needs, it might be worthwhile accepting 
a higher remaining ambiguity in favour of a higher recall.
\medskip

  The paper will be organized as follows: In section
\ref{seccio-bootstrapping} we will propose a bootstrapping 
procedure that combines the information of two taggers.
In section \ref{seccio-train0} we will describe the corpus used 
in the experiments, as well as the used analyzers, the
initial training set development and the initial results.
Section \ref{seccio-combinacio} is devoted to
describe the different experiments performed to find out 
the best way to combine the progressively obtained training 
corpora, and finally, in section \ref{seccio-best-tagger}, 
the best choice is presented and its results are reported. 
Preliminary work on extending the procedure to three voting 
taggers is discussed.

%
\section{Bootstrapping algorithm}
\label{seccio-bootstrapping}

The proposed bootstrapping algorithm is described in detail 
in figure~\ref{f-alg}. The meaning of the involved notation is 
described below:
\begin{itemize}
\item ${\cal C}^i$ stands for the retraining corpus of i-th iteration. 
In particular, ${\cal C}^0$ stands for the initial hand--tagged training 
corpus.
\item ${\cal T}$ stands for a hand--tagged test corpus used to estimate
the performance of the subsequent taggers.
\item ${\cal N}$ stands for the fresh part of the raw corpus used at 
each step to enlarge the training set. For simplicity we consider it 
independent of the specific iteration. 
\item $A_1$ and $A_2$ stand for both taggers (including, indistinctly,
the model acquisition and disambiguation algorithms).
\item $M_i^j$ stands for the j-th language model obtained by 
i-th tagger.
\item $train(A_i,{\cal C}^j)$ stands for the procedure of training the i-th 
tagger with the j-th training corpus. The result is the language model
$M_i^j$.
\item $test({\cal T},A_1,M_1^i,A_2,M_2^i)$ stands for a general procedure 
that returns the best accuracy obtained by any of the two taggers on the test set. 
\item $tag({\cal N},A_i,M_i^j)$ stands for the procedure of tagging the
raw corpus ${\cal N}$ with the i-th tagger using the j-th language 
model, producing ${\cal N}^i_j$.
\item $combine({\cal C}^0,{\cal N}_1^i\cap{\cal N}_2^i)$ is the general 
procedure of creation of (i+1)-th training corpus. This is done by
adding to the hand disambiguated corpus ${\cal C}^0$ the cases in 
${\cal N}$ in which both taggers coincide in their predictions (noted
${\cal N}_1^i\cap{\cal N}_2^i$).
\end{itemize}

\begin{figure}[htb]
\noindent\rule[0pt]{7.9cm}{0.4pt}
\vspace*{0mm}
{\small
{\tt\#\#\# Train taggers using manual corpus}\\
$M_1^0$ := $train(A_1,{\cal C}^0)$;\\
$M_2^0$ := $train(A_2,{\cal C}^0)$;\\
{\tt\#\#\# Compute achieved accuracy}\\
Acc-current := $test({\cal T},A_1,M_1^0,A_2,M_2^0)$;\\
Acc-previous := 0;\\
{\tt\#\#\# Initialize iteration counter}\\
$i$ := 0; \\
{\bf while} (Acc-current {\it significantly--better} Acc-previous) {\bf do}\\
\hspace*{3mm} ${\cal N}$ := {\it fresh--part--of--the--raw--corpus};\\ 
\hspace*{3mm} {\tt\#\#\# Tag the new data} \\
\hspace*{3mm} ${\cal N}_1^i$ := $tag({\cal N},A_1,M_1^i)$;\\
\hspace*{3mm} ${\cal N}_2^i$ := $tag({\cal N},A_2,M_2^i)$;\\
\vspace*{-1mm}\hspace*{3mm} {\tt\#\#\# Add the coincidence cases to}\\
\hspace*{3mm} {\tt\#\#\# the manual training corpus}\\
\hspace*{3mm} ${\cal C}^{i+1}$ := $combine({\cal C}^0,{\cal N}_1^i\cap{\cal N}_2^i)$;\\
\hspace*{3mm} {\tt\#\#\# retrain the taggers} \\
\hspace*{3mm} $M_1^{i+1}$ := $train(A_1,{\cal C}^{i+1})$;\\
\hspace*{3mm} $M_2^{i+1}$ := $train(A_2,{\cal C}^{i+1})$;\\
\hspace*{3mm} {\tt\#\#\# Prepare next iteration} \\
\hspace*{3mm} Acc-previous := Acc-current;\\
\hspace*{3mm} Acc-current := $test({\cal T},A_1,M_1^{i+1},A_2,M_2^{i+1})$\\
{\bf end-while}\\
}
\noindent\rule[0pt]{7.9cm}{0.4pt}\vspace*{-1.5mm}
\caption{{\small Bootstrapping algorithm using two taggers}}
\label{f-alg}
\end{figure}

In section~\ref{seccio-combinacio} we study the proper tuning of the 
algorithm in our particular domain by exploring the right size of the 
retrain corpus (i.e: the size of ${\cal N}$), the combination 
procedure (in particular we explore if a weighted combination is 
preferable to the simple addition) and the number of iterations that 
are useful. Additionally, we have tested if the (relatively cheap) process
of hand--correcting the disagreement cases between the two taggers 
at each step can give additional performance improvements.

%
\section{Tagging the {\sc LexEsp} Corpus}
\label{seccio-train0}

 
  The {\sc LexEsp} Project is a multi--disciplinary effort impulsed
by the Psychology Department from the University of Oviedo. It aims
to create a large database of language usage in order to enable
and encourage research activities in a wide range of fields, from 
linguistics to medicine, through psychology and artificial
intelligence, among others.

 One of the main issues of that database of linguistic resources
is the {\sc LexEsp} corpus, which contains 5.5 Mw of written
material, including general news, sports news, literature, 
scientific articles, etc.

  The corpus will be morphologically analyzed and disambiguated
as well as syntactically parsed. The used tagset is PAROLE compliant,
and consists of some 230 tags\footnote{There are potentially many more
                              possible tags, but they do not actually
                              occur.} 
fully expanded (using all information 
about gender, number, person, tense, etc.) which can be
reduced to 62 tags when only category and subcategory are considered.
\medskip

  The corpus has been morphologically analyzed with the {\sc maco+}
system, a fast, broad--coverage analyzer \cite{carmona98}.
The percentage of ambiguous words is 39.26\% and the average 
ambiguity ratio is 2.63 tags/word for the ambiguous words
(1.64 overall).
  The output produced by {\sc maco+}, is used as the input 
for two different POS taggers:
\begin{itemize}
\item {\sc Relax} \cite{padro96}. A relaxation--labelling based tagger 
which is able to incorporate information of different sources in a 
common language of weighted context constraints.
\item {\sc TreeTagger} \cite{marquez97b}. A decision--tree based
  tagger that uses a machine--learning supervised algorithm 
  for learning a base of statistical decision trees and an iterative
  disambiguation algorithm that applies these trees and filters out low
  probability tags.
\end{itemize}

  Since both taggers require training data, 96 Kw were hand
disambiguated\footnote{A trained human annotator can reach a rate of 
                       2000 words per hour, using a specially designed
                       Tcl/Tk graphical interface. So, 100Kw can be
                       annotated in about 50 man hours.} 
to get an initial training set (${\cal C}^0$) of 71 Kw and a 
test set ($\cal T$) of 25 Kw.

  The training set was used to extract bigram and trigram statistics
and to learn decision trees with {\sc TreeTagger}. The taggers also
require lexical probabilities, which were computed 
from the occurrences in the training corpus --applying smoothing 
(Laplace correction) to avoid zero probabilities--. For the words 
not appearing in the training set, the probability distribution for 
their ambiguity class was used. For unseen ambiguity classes, 
unigram probabilities were used.

  Initial experiments consisted of evaluating the precision of both
taggers when trained on the above conditions. Table
\ref{taula-train-manual} shows the results produced by each tagger.
The different kinds of information used by the relaxation labelling tagger
are coded as follows: {\sc B} stands for bigrams, {\sc T} for trigrams
and {\sc BT} for the joint set.
A baseline result produced by a most--frequent--tag tagger ({\sc Mft})
is also reported.

{\small\begin{table}[htb]
\begin{center}
\begin{tabular}{||l|c|c||} \hline
Tagger Model & Ambiguous & Overall\\\hline\hline
{\sc Mft}        & 88.9\% & 95.8\%\\
{\sc TreeTagger} & 92.1\% & 97.0\%\\
{\sc Relax (B)}   & 92.9\% & 97.3\%\\
{\sc Relax (T)}   & 92.7\% & 97.2\%\\
{\sc Relax (BT)}  & 93.1\% & 97.4\%\\\hline
\end{tabular}
\end{center}
\caption{Results of different taggers using the ${\cal C}^0$ training set}
\label{taula-train-manual}
\end{table}}

  These results point out that a 71 Kw training set manually
disambiguated provides enough evidence to allow the tagger
to get quite good results. Nevertheless, it is interesting to
notice that the trigram model has lower accuracy than the bigram
model. This is caused by the size of the training corpus, too small to
estimate a good trigram model.

%
\section{Improving Tagging Accuracy by Combining Taggers}
\label{seccio-combinacio}

In order to improve the model obtained from the initial hand tagged
training corpus, we may try a re--estimation procedure. The most
straightforward --and usual-- way of doing so is using a single 
tagger to disambiguate a fresh part of the corpus, and then use 
those data as a new training set. We will introduce the joint use 
of two taggers as a way to reduce the error rate introduced by the 
single tagger by selecting as retraining material only those cases
in which both taggers coincide. Two properties must hold for this 
method to work: 1) the accuracy in the cases of coincidence should be 
higher than those of both taggers individually considered, and 2)
the taggers should coincide in the majority of the cases 
(high coverage).

For instance, using a first set of 200Kw and given that both taggers 
agree in 97.5\% of the cases and that 98.4\% of those cases are
correctly tagged, we get a new corpus of 195Kw with an error rate
of 1.6\%. If we add the manually tagged 70Kw (assumed error free) 
from the initial training corpus we get a 265Kw corpus with an 
1.2\% error rate. 

\subsection{Size of the Retraining Corpus}

First of all, we need to establish which is the right size for the
fresh part of the corpus to be used as retraining data. We have
5.4Mw of raw data available to do so, but note that the bigger the 
corpus is, the higher the error rate in the retraining corpus will be
--because of the increasing proportion of new noisy corpus with respect
to the initial error free training corpus--.

So we will try to establish which is the corpus size at which further
enlargements of the retraining corpus don't provide significant
improvements.
Results for each tagger when retrained 
with different corpus sizes are shown in figure~\ref{fig-mides}
(accuracy figures are given over ambiguous words only). The
size at which both taggers produce their best result is that
of 800 Kw (namely ${\cal C}_{800}^1$), reaching 93.4\% and 93.9\%
accuracy on ambiguous words.

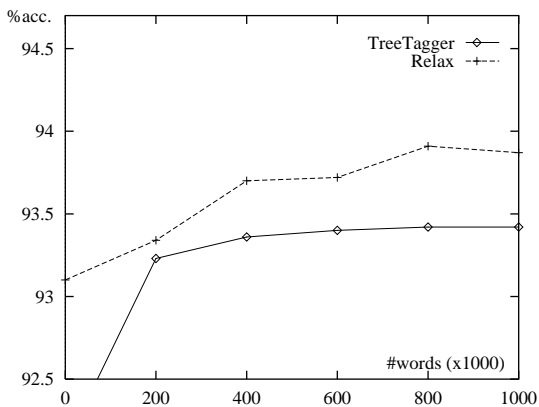
\figeps{mides}{Results using retraining sets of increasing sizes}{fig-mides}

The accuracies in figure~\ref{fig-mides} are computed retraining the
taggers with the coincidence cases in the retrain corpus, as described
in section \ref{seccio-bootstrapping}.

\subsection{Two taggers better than one}

 Once we have chosen a size for the retraining corpus, we will check
whether the joint use of two taggers to reduce the error in the
training corpus is actually better than retraining only with a 
single tagger.

{\small\begin{table}[htb]
\begin{center}
\begin{tabular}{||l|l|l||} \hline
Tagger Model     & single  & ${\cal C}_{800}^1$  \\\hline\hline
{\sc TreeTagger} & 93.0\%  & 93.4\%   \\
{\sc Relax (BT)} & 93.7\%  & 93.9\%   \\ \hline
\end{tabular}
\end{center}
\caption{Comparative results when retraining with a new 800Kw corpus}
\label{taula-1-sol-tagger}
\end{table}}

  Comparative results obtained for each of our taggers when using
retraining material generated by a single tagger (the size of the fresh part
of the corpus to be used as retrain data was also 800 Kw) and when using
${\cal C}_{800}^1$ are reported in table \ref{taula-1-sol-tagger}. Those
results point that the use of two taggers to generate the retraining
corpus, slightly increases the accuracy of any tagger since it
provides a less noisy model.

  The error rate in the retrain corpus when using the {\sc Relax-BT} tagger alone
is 2.4\%, while when using the coincidences of both taggers is reduced
to 1.3\%. This improvement in the training corpus quality enables the
taggers to learn better models and slightly improve their performance.
 Probably, the cause that the performance improvement is not larger
must not be sought in the training corpus error rate, but in the
learning abilities of the taggers.

\subsection{Number of Iterations}

  The bootstrapping algorithm must be stopped when no further
improvements are obtained. This seems to happen after the first 
iteration step. Using the 800Kw from the beginning yields
similar results than progressively enlarging the corpus size at each step.
Results are shown in table~\ref{taula-ret2}. 
 
{\small\begin{table}[htb]
\begin{center}
\begin{tabular}{||l|l|l||} \hline
Tagger Model      & ${\cal C}_{800}^1$ & ${\cal C}_{800}^2$  \\\hline\hline
{\sc TreeTagger}  & 93.4\%  & 93.5\% \\
{\sc Relax (BT)}  & 93.9\%  & 93.8\%\\ \hline
\end{tabular}
\end{center}
\caption{Results when retraining with a 800Kw corpus in one and two steps}
\label{taula-ret2}
\end{table}}

  Facts that support this conclusion are:
\begin{itemize}
\item The variations with respect to the results for one re--estimation 
      iteration are not significant.
\item {\sc TreeTagger} gets a slight improvement while {\sc Relax} 
      decreases ---indicating that the re--estimated model does not 
      provide a clear improvement--.
\item The intersection corpora used to retrain have the same accuracy
      (98.4\%) both in iteration one and two, and the difference in the number or
      coincidences (97.7\% in iteration one vs. 98.3\% in iteration two) is
      not large enough to provide extra information.
\end{itemize}




\subsection{Use of Weighted Examples}

  We have described so far how to combine the results of two POS 
taggers to obtain larger training corpora with low error rates.
We have also combined the agreement cases of both taggers with the
initial hand--disambiguated corpus, in order to obtain a less
noisy training set. Since the hand--disambiguated corpus offers a
higher reliability than the tagger coincidence set, we might want 
to establish a {\sl reliability} degree for our corpus, by means of 
controlling the contribution of each part. This can be done through the
estimation of the error rate of each corpus, and establishing
a weighted combination which produces a new retraining corpus with 
the desired error rate.

  As mentioned above, if we put together a 
hand--disambiguated (assumed error-free) 70Kw corpus and 
a 195Kw automatically tagged corpus with an estimated
error rate of 1.6\%, we get a 265Kw corpus with a
1.2\% error rate. But if we combine them with different weights
we can control the error rate of the corpus:
e.g. taking the weight for the correct 70Kw twice the
weight for the 195Kw part, we get a corpus of
335Kw\footnote{Obviously this occurrences are {\sl virtual}  
               since part of them are duplicated.} 
with an error rate of 0.9\%. In that way we can adjust 
the weights to get a training corpus with the desired error rate.

  Figure~\ref{fig-err} shows the relationship between the
error rate and the relative weights between ${\cal C}^0$ 
and the retraining corpus.

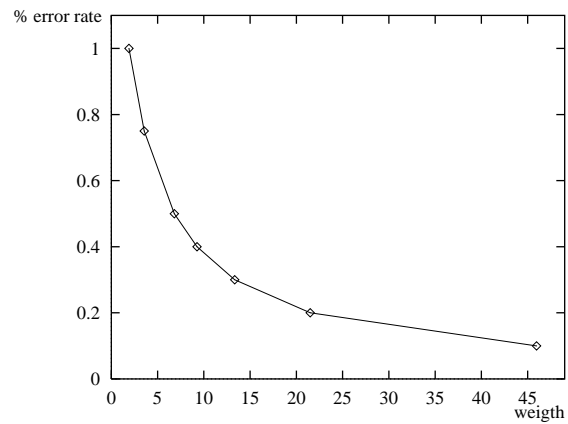
\figeps{pes-err}{Relationship between the error rate 
and the relative weights for training corpora.}{fig-err}


 This weighted combination enables us to dim the undesired 
effect of noise introduced in the automatically tagged part of the 
corpus.

  This combination works as a kind of back-off interpolation between
the correct examples of ${\cal C}^0$ and the slightly noisy corpus of
coincidences added at each step.
By giving higher weights to the former, cases well represented in 
the ${\cal C}^0$ corpus are not seriously influenced by new erroneous 
instances, but cases not present in the ${\cal C}^0$ corpus are still
incorporated to the model.
 So, the estimations of the statistical parameters for ``new'' cases
will improve the tagging performance while statistical estimations 
of already well represented cases will be, at most, slightly poorer. 

We have performed an experiment to determine the performance
obtained when the taggers are trained with corpus obtained combining
${\cal C}^0$ and the first extension of 200,000 words
(${\cal N}_1^1\cap{\cal N}_2^1$)  with different relative 
weights\footnote{Weights were straightforwardly incorporated to 
                 the bigrams and trigrams statistics. The decision 
                 tree learning algorithm had to be slightly 
                 modified to deal with weighted examples.}.
 The steps selected are the weights corresponding to error rates
of 0.1\%, 0.2\%, 0.3\%, 0.4\%, 0.5\%, 0.75\% and 1\%. 

It is obvious that too high weighting in favour of initial examples will 
produce a lower error rate (tending to zero, the same than the manual
corpus), but it will also bias the taggers to behave like the initial
tagger, and thus will not take advantage of the new cases.

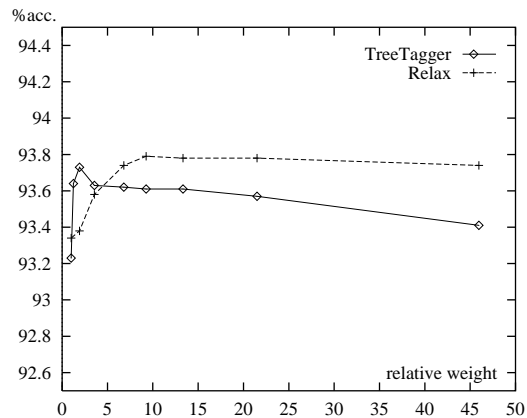
\figeps{pesos}{Results using the ${\cal C}_{200}^1$ training set with different weightings}{fig-pesos}

The results summarized in figure~\ref{fig-pesos} show that 
there is a clear improvement in the tagger
performance when combining two training corpora with a proper 
weighting adjustment. Obviously, there is a tradeoff point where
the performance starts to decrease due to an excessive weight for the
initial data. 

Although the behaviour of both curves is similar, it is also clear
that the different tagging algorithms are not equally sensitive to 
the weighting values: In particular, {\sc TreeTagger} achieves its 
highest performance for weights between 1 and 3, while {\sc Relax-BT}
needs a weight around 10. 



\subsection{Hand--correcting Disagreement Cases}

 Another possible way to reduce the error rate in the training corpus
is hand correcting the disagreement cases between taggers. This
reduces the error rate of the new training corpus at a low human
labor cost, since the disagreement cases are only a small part of the
total amount.

  For instance, In ${\cal C}_{200}^1$ corpus, there were 5,000 disagreement
cases. Hand--correcting and adding them to the previous set we
obtain a slightly larger corpus\footnote{The increasing in number of 
training examples is specially noticeable in the case of decision
trees (+14,000). This is due to the fact that each example considers a
context window of six items. After hand--correction all sequences of six
words are valid while before correction it was quite probable to find gaps
(cases of disagreement) in the sequences of six words of the
intersection corpus.} (270Kw) with a slightly lower error
rate (1.17\%), which can be used to retrain the taggers. We call this 
corpus ${\cal C}_{\rm M}^1$ (M stands for manual revision).

  Results obtained with the corrected retraining corpus are shown in 
table~\ref{taula-manual}, together with the results obtained with
fully automatic retraining corpus of 200 Kw (${\cal C}_{200}^1$) and 
800 Kw (${\cal C}_{800}^1$).

{\small\begin{table}[htb]
\begin{center}
\begin{tabular}{||l|l|l|l||} \hline
Tagger Model& ${\cal C}_{200}^1$ & ${\cal C}_{\rm M}^1$ & ${\cal C}_{800}^1$\\\hline\hline
{\sc TreeTagger}  & 93.2\% & 93.8\% & 93.4\%\\
{\sc Relax (BT)}  & 93.3\% & 93.8\% & 93.9\%\\\hline
\end{tabular}
\end{center}
\caption{Comparative results using ${\cal C}_{200}^1$, ${\cal C}_{\rm
M}^1$ and ${\cal C}_{800}^1$ training sets}
\label{taula-manual}
\end{table}}

  The first conclusion in this case is that the hand--correction of
disagreement cases gives a significant accuracy improvement in
both cases.  However, the gain obtained is the same order than that
obtained with a larger retraining corpus automatically disambiguated.
Unfortunately we had neither more available human resources nor time to hand-correct
the remaining 15,000 disagreement words of ${\cal C}_{800}^1$ in order
to test if some additional improvement can be achieved from the best
automatic case. Without performing this experiment it is impossible to
extract any reliable conclusion. However, we know that the price to
pay for an uncertain accuracy gain is the effort of manually tagging 
about 20,000 words. Even when that would mean an improvement, we suspect
that it would be more productive to spend this effort in constructing 
a larger initial training corpus.

  Thus, unless there is a very severe restriction on the size of the
available retraining corpus, it seems to be cheaper and faster not 
to hand correct the disagreement cases.
%
\section{Best Tagger}
\label{seccio-best-tagger}

   All the the above described combinations produce a wide range of 
possibilities to build a retraining corpus. We can use retraining
corpus of different sizes, perform several retraining steps, and
weight the combination of the retraining parts. Although all 
possible combinations have not been explored, we 
have set the basis for a deeper analysis of the possibilities.

  A promising combination is using the more reliable information 
obtained so far to build a ${\cal C}^1_{\rm Best}$ retraining corpus, 
consisting of ${\cal C}^1_{\rm M}$ (which includes ${\cal C}^0$) plus the
coincidence cases from the ${\cal C}^0_{800}$ which were not included
in ${\cal C}^1_{\rm M}$. This combination has only been tested in its
straightforward form, but we feel that the weighted combination of the
constituents of ${\cal C}^1_{\rm Best}$ should produce better
results than the reported so far.

   On the other hand, the above reported results were obtained using only
either the {\sc TreeTagger} with decision trees information or the
{\sc Relax} tagger using bigrams and/or trigrams. Since the 
{\sc Relax} tagger is able to combine different kinds of constraints,
we can write the decision trees learned by {\sc TreeTagger} in the
form of constraints ({\sc C}), and make {\sc Relax} use them as in \cite{marquez97a}.

Table~\ref{taula-millor} shows the best results obtained with every
combination of constraint kinds. The retraining corpora which
yield each result are also reported.

{\small\begin{table}[htb]
\begin{center}
\begin{tabular}{||l|l|l|l||} \hline
Tagger Model&Amb.&Overall&Corpus\\\hline\hline
{\sc TreeTagger}  & 93.8\% & 97.7\%& ${\cal C}^1_{\rm M}$\\
{\sc Relax (B)}   & 93.3\% & 97.5\%& ${\cal C}^1_{\rm Best}$\\
{\sc Relax (T)}   & 93.7\% & 97.6\%& ${\cal C}^1_{\rm Best}$\\
{\sc Relax (BT)}  & 93.9\% & 97.7\%& ${\cal C}^1_{800}$ ${\cal C}^1_{1000}$\\
{\sc Relax (C)}   & 93.8\% & 97.7\%& ${\cal C}^1_{\rm Best}$\\
{\sc Relax (BC)}  & 94.1\% & 97.8\%& ${\cal C}^1_{200}$ ${\cal C}^1_{\rm M}$ ${\cal C}^1_{\rm Best}$\\
{\sc Relax (TC)}  & 94.2\% & 97.8\%& ${\cal C}^1_{200}$\\
{\sc Relax (BTC)} & 94.2\% & 97.8\%& ${\cal C}^1_{400}$ ${\cal C}^1_{\rm Best}$\\\hline
\end{tabular}
\end{center}
\caption{Best results for each tagger with all possible constraint combinations}
\label{taula-millor}
\end{table}}

  Further experiments must establish which is the most appropriate
bootstrapping policy, and whether it depends on the used taggers.

\section{Conclusions and Future Work}
\label{seccio-conclusions}


   We presented the collaboration between two different POS taggers in a 
voting approach as a way to increase tagging accuracy. Since words
in which both taggers make the same prediction present a higher
accuracy ratio, tagger collaboration can also be used to develop large
training sets with a low error rate, which is specially useful for 
languages with a reduced amount of available linguistic resources.

  The presented results show that:
\begin{itemize}
\vspace*{-0.1cm}
\item The precision of the taggers 
      taking into account only the cases in which they agree, is
      significantly higher than overall cases. 
      Although this is not useful to disambiguate a 
      corpus, it may be used in new corpus development to reduce the amount 
      of hand tagging while keeping the noise to a minimum. This has been 
      used in a bootstrapping procedure to develop an annotated corpus of 
      Spanish of over 5Mw, with an estimated accuracy of 97.8\%.
\vspace*{-0.1cm}
\item Obviously, the recall obtained joining the proposals from both 
      taggers is higher than the results of any of them separately and a 
      remaining ambiguity is introduced, which causes a decrease in
      precision. Depending on the user needs, it might be worthwhile 
      accepting a higher remaining ambiguity in favour of a higher
      recall. With the models acquired from the best training corpus,
      we get a tagger with a recall of 98.3\% and a remaining
      ambiguity of 1.009 tags/word, that is, 99.1\% of the words are 
      fully disambiguated and the remaining 0.9\% keep only two tags.
\end{itemize}
     
   This procedure can easily be extended to a larger number of
taggers. We are currently studying the collaboration of three taggers,
using a ECGI tagger \cite{pla98} in addition to the other two.
Preliminary results point that the cases in which the three taggers 
coincide, present a higher accuracy than when only two taggers are
used (96.7\% compared to 95.5\% on ambiguous words) and that the 
coverage is still very high (96.2\% compared to 97.7\%).
  
Nevertheless, the difference is relatively small, and it must be further
checked to establish whether it is worth using a larger number of
taggers for building low error rate training corpora.  In addition, as
pointed in \cite{padro98b}, the error in test corpora may introduce
distortion in the evaluation and invalidate small improvements and
although we have used a manually disambiguated test corpus, it may
contain human errors. For all this reasons, much work on improving the test
corpus and on validating the so far obtained results is still to be done.

\section{Acknowledgments}

This research has been partially funded by the Spanish Research
Department (CICYT's ITEM project TIC96--1243--C03--02), by the 
EU Commission (EuroWordNet LE4003) and by the Catalan Research 
Department (CIRIT's quality research group 1995SGR 00566).


\end{document}